%% file: main.tex
\documentclass[10pt,twocolumn,letterpaper]{article}

\usepackage[pagenumbers]{cvpr}

\input{preamble}

\definecolor{cvprblue}{rgb}{0.21,0.49,0.74}
\usepackage[pagebackref=false,breaklinks,colorlinks,allcolors=cvprblue]{hyperref}

\usepackage[accsupp]{axessibility}

\usepackage{algorithm}
\usepackage{algpseudocode}
\usepackage{amsmath}
\usepackage{amsfonts}
\usepackage{amssymb}
\usepackage{array}
\usepackage{bm}
\usepackage{booktabs}
\usepackage{colortbl}
\usepackage{dsfont}
\usepackage{fontawesome}
\usepackage{makecell}
\usepackage{multirow}
\usepackage{pifont}
\usepackage{tikz}

\definecolor{my_green}{RGB}{51,102,0}
\definecolor{my_red}{RGB}{204, 0, 0}
\definecolor{merit_red}{RGB}{255,46,99}
\definecolor{mgt_red}{RGB}{255,46,99}
\definecolor{mgt_dark}{RGB}{37,42,52}
\definecolor{mgt_blue}{RGB}{8,217,214}
\definecolor{mgt_gray}{RGB}{156,156,156}

\newcommand{\name}{\textbf{\textcolor{mgt_dark}{Edit}\textcolor{mgt_red}{M}\textcolor{mgt_gray}{G}\textcolor{mgt_blue}{T}}}

\newcommand{\dataname}{CrispEdit-$2$M}

\definecolor{link}{RGB}{156,156,156}
\hypersetup{
  colorlinks=true,
  linkcolor=link,
  citecolor=link,
  urlcolor=link,
}

\def\paperID{}
\def\confName{CVPR}
\def\confYear{2026}

\title{Masked Generative Transformer Is What You Need for Image Editing}

\author{Wei~Chow$^{1,2,}$\footnotemark[1] \quad Linfeng~Li$^{1,}$\footnotemark[1] \quad Xian Sun$^3$ \quad Lingdong~Kong$^{1,2}$ \quad Zefeng~Li$^1$ \quad Qi~Xu$^1$ \\
Hang~Song$^1$ \quad Tian~Ye$^5$ \quad Xian~Wang$^1$ \quad Jinbin~Bai$^2$ \quad Shilin~Xu$^1$ \quad Xiangtai~Li$^1$ \\
Junting~Pan$^1$ \quad Shaoteng~Liu$^1$ \quad Ran~Zhou$^1$ \quad Tianshu~Yang$^1$ \quad Songhua~Liu$^{4,}$\footnotemark[2] \\
\small $^1$ByteDance\quad $^2$National University of Singapore\quad
\small $^3$Duke University\quad
\small $^4$Shanghai Jiao Tong University\quad $^5$HKUST(GZ)
\\[1ex]
{\faGlobe}~\textbf{\textsl{Project Page, Code \& Dataset:}}~\href{https://weichow23.github.io/EditMGT/}{\textcolor{mgt_red}{\textsl{https://weichow23.github.io/EditMGT}}}
}

\begin{document}

\twocolumn[{
    \vspace{-0.45cm}
    \renewcommand\twocolumn[1][]{#1}
    \maketitle
    \begin{center}
    \centering
    \captionsetup{type=figure}
    \vspace{-0.7cm}
    \includegraphics[width=\textwidth]{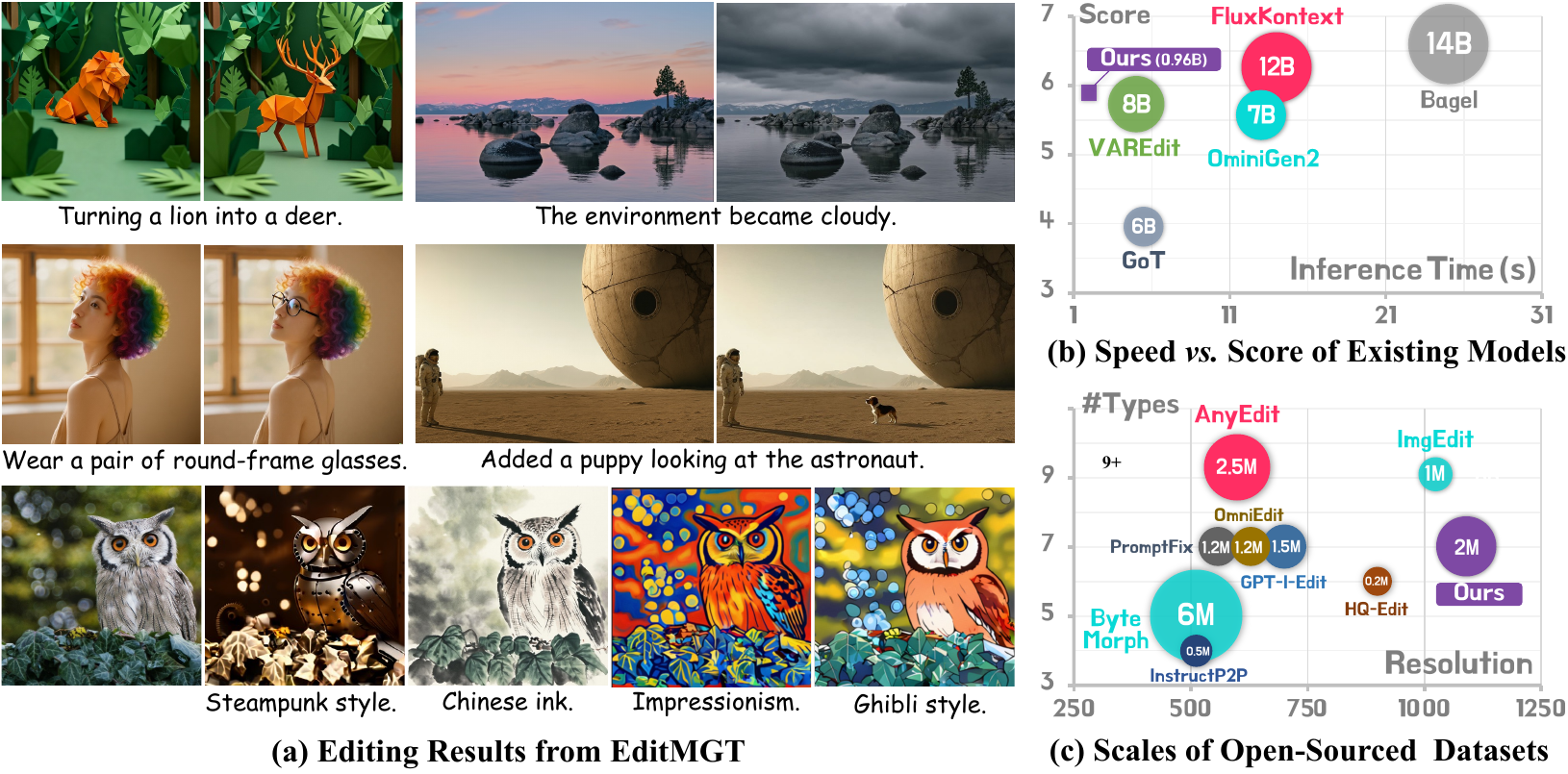}
    \vspace{-0.65cm}
    \caption{\textbf{Overview of \name{} and \dataname{}.} We introduce the first MGT-based editing model that performs editing in $2$s with $960$M parameters, $6\times$ faster than existing models of comparable performance while surpassing $8$B models. We also contribute \dataname{}, providing $2$M high-resolution ($\geq$$1024$) editing samples across $7$ categories.}
    \label{fig:teaser}
    \vspace{0.1cm}
    \end{center}
}]

\input{sec/0_abstract}
\input{sec/1_intro}
\input{sec/2_method}
\input{sec/3_experiments}
\input{sec/4_conclusion}

{
    \small
    \bibliographystyle{ieeenat_fullname}
    \bibliography{main}
}

\clearpage\clearpage
\begin{figure*}[t]
    \centering
    \includegraphics[width=\linewidth]{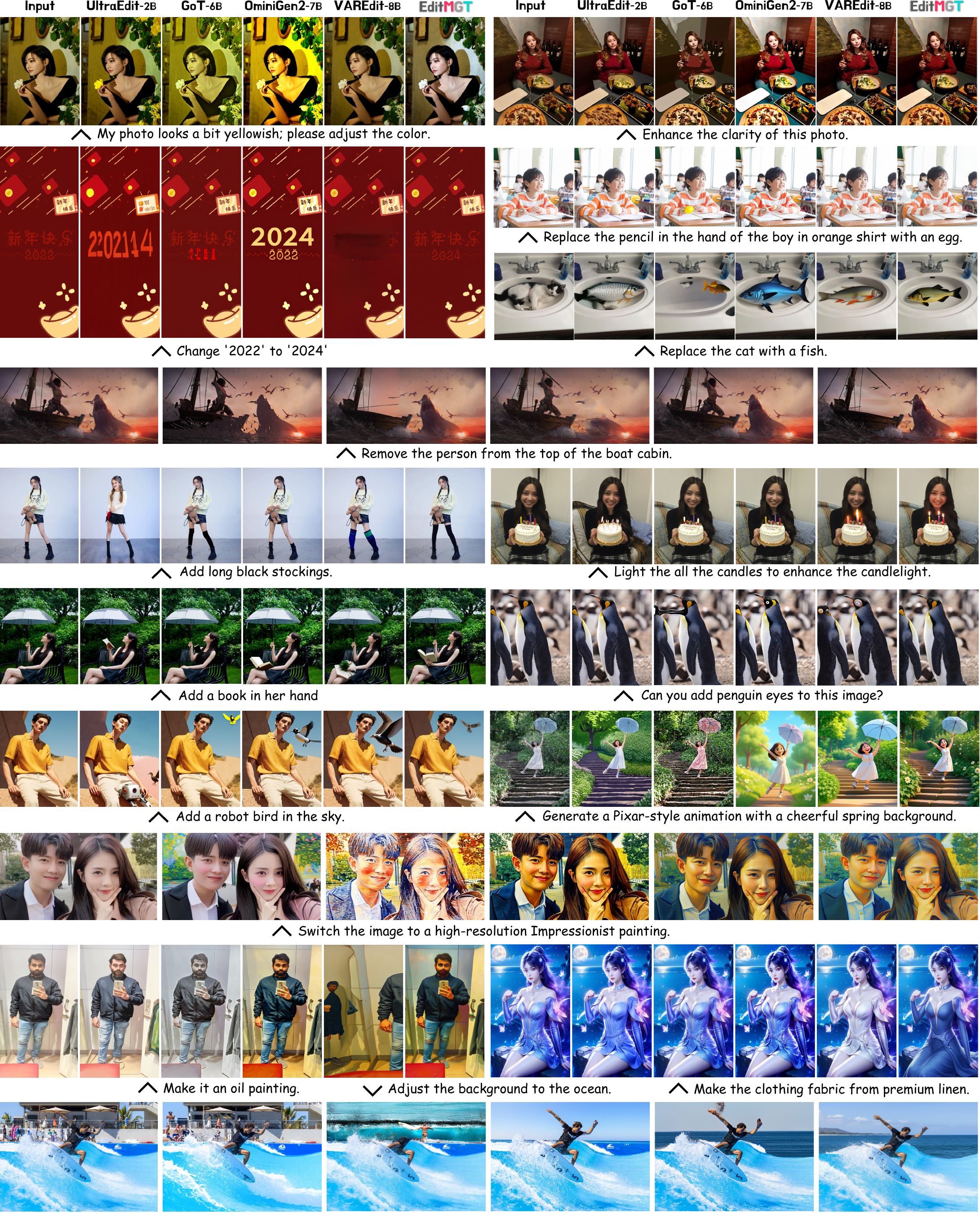}
    \vspace{-0.5cm}
    \caption{\textbf{Qualitative comparisons.} \name{} (960M) outperforms larger models across diverse editing tasks such as object transformation, scene replacement, and material substitution.}
    \label{fig:quality}
\end{figure*}

\end{document}

%% file: preamble.tex
\usepackage{xspace}

\makeatletter
\DeclareRobustCommand\onedot{\futurelet\@let@token\@onedot}
\def\@onedot{\ifx\@let@token.\else.\null\fi\xspace}

\def\ie{\textit{i.e}\onedot}

\makeatother

%% file: sec/0_abstract.tex
\begin{abstract}
Diffusion models dominate image editing, yet their global denoising mechanism entangles edited regions with surrounding context, causing modifications to propagate into areas that should remain intact. We propose a fundamentally different approach by leveraging Masked Generative Transformers (MGTs), whose localized token-prediction paradigm naturally confines changes to intended regions. We present \name{}, an MGT-based editing framework that is the first of its kind. Our approach employs \textit{multi-layer attention consolidation} to aggregate cross-attention maps into precise edit localization signals, and \textit{region-hold sampling} to explicitly prevent token flipping in non-target areas. To support training, we construct \dataname{}, a $2$M-sample high-resolution ($\geq$1024) editing dataset spanning seven categories. With only $960$M parameters, \name{} achieves state-of-the-art image similarity on multiple benchmarks while delivering $6\times$ faster editing, demonstrating that MGTs offer a compelling alternative to diffusion-based editing.
\end{abstract}
\vspace{-0.5cm}

%% file: sec/1_intro.tex
\section{Introduction}
\label{sec:intro}

\begin{figure*}[t]
    \centering
    \includegraphics[width=0.98\linewidth]{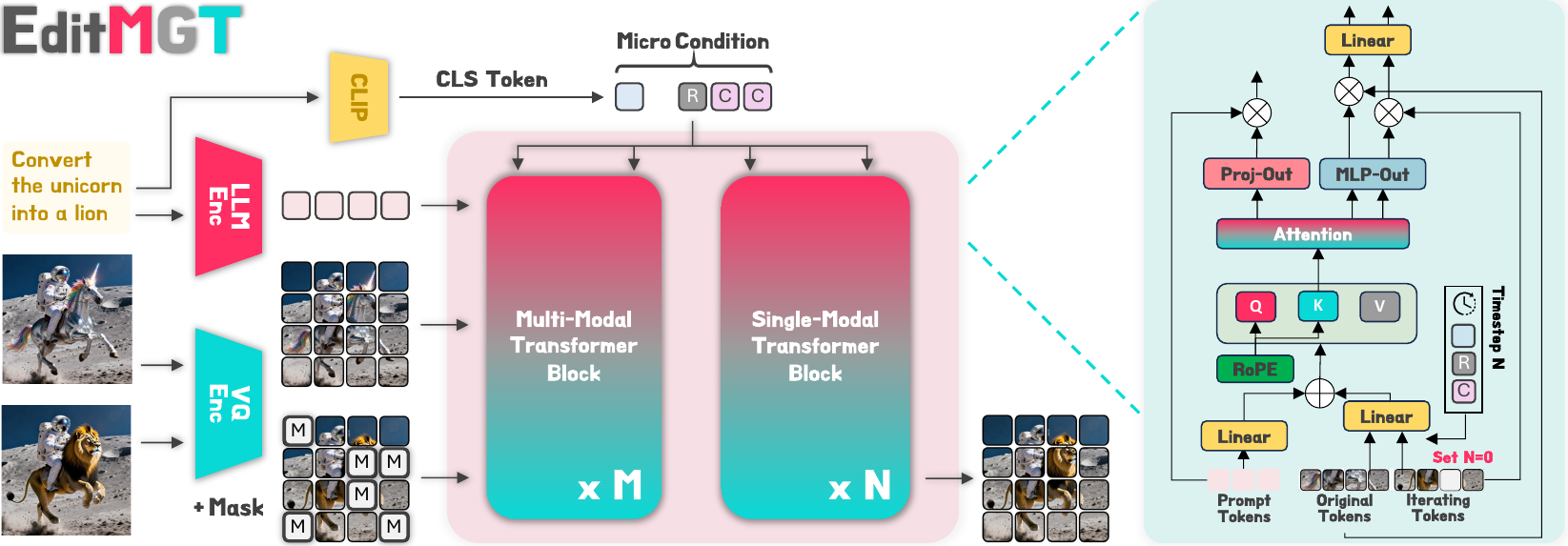}
    \vspace{-0.3cm}
    \caption{\textbf{\name{} framework.} The original image conditions generation via attention injection. Right panel: token interactions inside the multi-modal and single-modal transformer blocks.}
    \label{fig:method}
    \vspace{-0.2cm}
\end{figure*}

Instruct a diffusion model to \emph{``put a birthday hat on the dog''}, and you may find the dog's fur subtly recolored, the background shifted, or the lighting altered in ways you never requested. This \emph{edit leakage} stems from the global denoising dynamics of diffusion models (DMs)~\cite{hertz2022prompt}: every step refines the entire image simultaneously, making it inherently difficult to confine modifications to a specific region. Prior solutions based on large-scale training~\cite{yu2025anyedit}, predefined masks~\cite{zhang2024magicbrush}, or inversion techniques~\cite{mokady2023null} mitigate but do not fully resolve this entanglement.

We observe that a different class of generative models sidesteps this problem entirely. Masked Generative Transformers (MGTs)~\cite{chang2022maskgit} synthesize images by iteratively predicting masked tokens in parallel rather than performing holistic refinement. This localized decoding paradigm naturally supports selective token replacement: tokens corresponding to non-target regions can simply be held fixed, providing an architectural guarantee against edit leakage that DMs lack. The question here is whether this token-level control can be harnessed for high-quality, instruction-driven image editing.

We answer affirmatively with \name{}~\cite{chow2026editmgt}, an MGT-based editing framework that is, to our knowledge, the first of its kind. We aim to address two essential capabilities for effective editing. First, for \emph{adaptive localization}, we find that MGT cross-attention maps encode informative semantic correspondences between text instructions and visual regions, but lack sufficient prominence for direct use. We devise \textit{multi-layer attention consolidation} that aggregates and refines these maps across network layers, producing sharp, accurate localization of edit-relevant areas (Fig.~\ref{fig:attention}). Second, for \emph{region preservation}, we propose \textit{region-hold sampling}, a strategy that restricts token updates within low-attention areas during iterative decoding, explicitly preventing modifications from propagating into non-target regions.
To train EditMGT, we assemble \dataname{}, a high-resolution ($\geq$1024) editing dataset with $2$M rigorously filtered samples across seven distinct categories. We adapt a pretrained text-to-image MGT (Meissonic~\cite{bai2024meissonic}) into an editing model via attention injection, requiring no additional parameters.

%% file: sec/2_method.tex
\section{Methodology}
\label{sec:method}

\begin{figure*}[t]
    \centering
    \includegraphics[width=\linewidth]{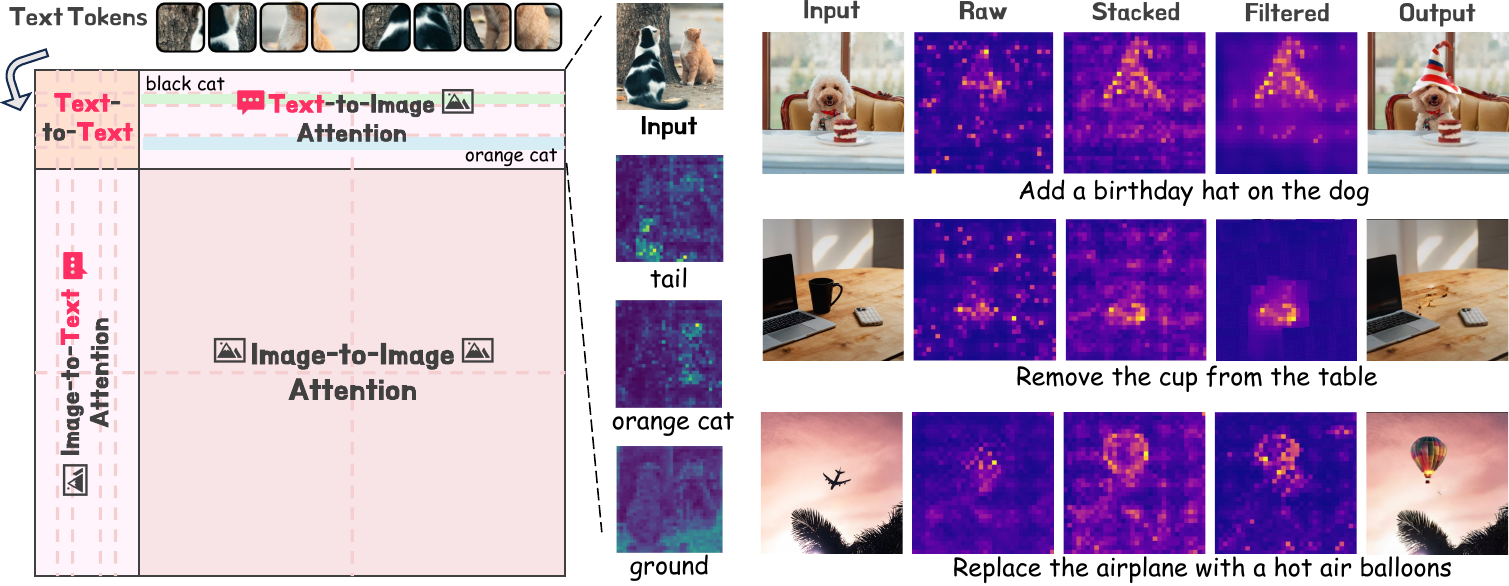}
    \vspace{-0.6cm}
    \caption{\textbf{Attention mechanism in \name{}.} Cross-attention maps encode semantic correspondences between instructions and visual regions. Multi-layer consolidation sharpens these maps for region-hold sampling.}
    \label{fig:attention}
    \vspace{-0.2cm}
\end{figure*}

\noindent\textbf{Preliminary: Masked Generative Transformers.}
MGTs synthesize images by starting from a fully masked token grid and progressively revealing tokens via parallel prediction and confidence-based resampling~\cite{chang2022maskgit}. Let $C_I \in \mathbb{R}^{N \times d}$ and $C_T \in \mathbb{R}^{M \times d}$ denote image and text tokens. Building upon Meissonic~\cite{bai2024meissonic}, each transformer block applies RoPE~\cite{su2024roformer} for spatial encoding, then computes multi-modal attention over $C = [C_I; C_T]$:
$\textbf{W} = \text{softmax}({QK^{\top}}/{\sqrt{d}})$.
This token-level control makes MGTs naturally suited for editing: tokens in non-target regions can remain fixed throughout decoding.

\noindent\textbf{Image-Conditioned Editing via Attention Injection.}
To condition the generation on the original image, we introduce image condition tokens $C_V \in \mathbb{R}^{N \times d}$ with the same shape as $C_I$ (Fig.~\ref{fig:method}). Critically, $C_V$ shares all parameters with $C_I$ and uses identical RoPE positions, \ie, $(i,j)_{C_V} = (i,j)_{C_I}$, ensuring precise spatial alignment between the source and edited images. The key distinction is that the timestep assigned to $C_V$ is held constant at zero, preventing temporal drift and maintaining a stable reference signal.

The model $\theta$ is trained by minimizing the cross-entropy loss over masked token reconstruction, jointly conditioned on visible tokens and the image reference:
\begin{equation}
L = \mathbb{E}_{(x, t) \sim \mathcal{D},\, \mathbf{m} \sim \mathcal{M}}
\bigg[- \sum\nolimits_{i \in \mathbf{m}} \log p_\theta(v_i \mid v_{\neg i}, C_T; C_V)\bigg],
\end{equation}
where $\mathbf{m}$ is a binary mask selecting tokens to predict, $v_{\neg i}$ denotes visible tokens, and the masking rate follows a cosine schedule sampled from a truncated $\arccos$ distribution. To control $C_V$'s influence at inference, we introduce a bias matrix $\mathcal{E}$ into the attention weights as $\textbf{W}_{\text{new}} = \textbf{W} + \mathcal{E}$, where $\mathcal{E}$ scales attention between $C_I$ and $C_V$ by $\log(\gamma)$ while leaving intra-modality patterns unchanged. Setting $\gamma = 0$ removes conditioning; $\gamma > 1$ strengthens it. This mechanism transforms the pretrained text-to-image MGT into an editor via attention injection alone, without introducing any additional parameters.

\input{table/edit_benchmark}

\noindent\textbf{Multi-Layer Attention Consolidation.}
We find that the cross-attention weights between instruction tokens $C_T$ and image tokens $C_I$ naturally encode the spatial location of intended edits (Fig.~\ref{fig:attention}). Notably, even within the initial iterations, the model can delineate the contours of objects to be added or modified, establishing precise text-to-image correspondence. However, individual layer activations lack sufficient prominence and exhibit unclear boundaries with internal discontinuities, making them unreliable for direct use as localization masks.

To remedy this, we aggregate attention from the most coherent single-modality processing layers (blocks 28--36), which we empirically identify as producing the most semantically aligned activations. The aggregated map is further refined via adaptive filtering~\cite{diniz1997adaptive} to suppress spatial discontinuities and sharpen object boundaries. The resulting consolidated map provides a reliable per-token localization signal that distinguishes edit-relevant regions from those that should remain intact.

\noindent\textbf{Region-Hold Sampling.}
Given the consolidated attention map, we compute a localization score for each image token:
\begin{equation}
    s_L = \frac{1}{|\mathcal{L}||\mathcal{M}|}\sum\nolimits_{\ell \in \mathcal{L},\, m \in \mathcal{M}} \mathcal{W}^{\ell}_i[m,:] \;\in \mathbb{R}^N,
\end{equation}
where $\mathcal{W}^{\ell}_i \in \mathbb{R}^{M \times N}$ is the normalized instruction-to-image attention at layer $\ell$, $\mathcal{L}$ is the set of selected layers, and $\mathcal{M}$ indexes the relevant instruction tokens (which can be restricted to specific keywords for finer control). During iterative decoding, the MGT reveals high-confidence tokens while re-masking uncertain ones. With region-hold sampling, tokens satisfying $s_L < \lambda$ are additionally restored to their original values from $C_V$ after each step, where $\lambda$ is a user-controllable threshold that trades off edit extent against preservation.

\noindent\textbf{Training Pipeline.}
We assemble \dataname{} comprising $2$M high-resolution ($\geq$1024) editing samples across seven categories with rigorous quality filtering. Training proceeds in three stages: (1) base model adaptation with Gemma2-2B~\cite{team2024gemma2} as text encoder ($5$K steps), (2) full fine-tuning on $4$M editing samples ($50$K steps), and (3) high-quality refinement for human preference alignment ($1$K steps). Additional details are available in the main paper~\cite{chow2026editmgt}.

%% file: table/edit_benchmark.tex
\begin{table*}[t]
    \centering
    \caption{\textbf{Comparative results} for instructive image editing on the test sets of EMU Edit~\cite{sheynin2024emu} and MagicBrush~\cite{zhang2024magicbrush}.
    We list the task-specific models in the first block and some concurrent universal models in the second block.}
    \vspace{-0.2cm}
    \label{tab:eval_image_editing}
    \resizebox{0.95\textwidth}{!}{
        \centering
        \begin{tabular}{l|r|c|cccc|cccc}
            \toprule
            & & & \multicolumn{4}{c|}{\textbf{EMU Edit Test Set}} & \multicolumn{4}{c}{\textbf{MagicBrush Test Set}}
            \\
            \textbf{Model} & \textbf{Venue} & \textbf{Size} & $\textbf{CLIP}_\mathrm{im}\!\uparrow$ & $\textbf{CLIP}_\mathrm{out}\!\uparrow$ &  ~~~$\textbf{L1}\!\downarrow$~~~  & ~\textbf{DINO}$\uparrow$~  &  $\textbf{CLIP}_\mathrm{im}\!\uparrow$ & $\textbf{CLIP}_\mathrm{out}\!\uparrow$ &  ~~~$\textbf{L1}\!\downarrow$~~~ & ~\textbf{DINO}$\uparrow$~
            \\
            \midrule
            \midrule
            \textcolor{mgt_gray}{$\bullet$}~InstructPix2Pix~\cite{brooks2023instructpix2pix} & CVPR'23 & $1$B & $0.834$ & $0.219$ & $0.121$ & $0.762$ & $0.837$ & $0.245$ & $0.093$ & $0.767$ 
            \\
            \textcolor{mgt_gray}{$\bullet$}~MagicBrush~\cite{zhang2024magicbrush} & NeurIPS'23 & $1$B & $0.838$ & $0.222$ & $0.100$ & $0.776$ & $0.883$ & $0.261$ & $0.058$ & $0.871$ 
            \\
            \textcolor{mgt_gray}{$\bullet$}~PnP~\cite{tumanyan2023plug} & CVPR'23  & $1$B  & $0.521$ & $0.089$ & $0.304$ & $0.153$ & $0.568$ & $0.101$ & $0.289$ & $0.220$ 
            \\
            \textcolor{mgt_gray}{$\bullet$}~Null-Text Inv.~\cite{mokady2023null} & CVPR'23 & $1$B & $0.761$ & $0.236$ & $0.075$ & $0.678$ & $0.752$ & $0.263$ & $0.077$ & $0.664$ 
            \\
            \textcolor{mgt_gray}{$\bullet$}~UltraEdit~\cite{zhao2024ultraedit} & NeurIPS'24 & $2$B & $0.793$ & $0.283$ & $0.071$ & $\mathbf{0.844}$ & $0.868$ & -  & $0.088$ & $0.792$ 
            \\   
            \textcolor{mgt_gray}{$\bullet$}~EMU Edit~\cite{sheynin2024emu} & CVPR'24 & - & $0.859$ & $0.231$ & $0.094$ & $0.819$ & $0.897$ & $0.261$ & \underline{$0.052$} & \underline{$0.879$} 
            \\
            \textcolor{mgt_gray}{$\bullet$}~AnyEdit~\cite{yu2025anyedit} & CVPR'25 & $1$B & $0.872$ & $0.285$ & \underline{$0.070$} & $0.821$ &  $0.898$ & $0.275$ & $\mathbf{0.051}$ & $0.881$
            \\
            \midrule
            \textcolor{mgt_blue}{$\bullet$}~OmniGen~\cite{xiao2025omnigen} & arXiv'24 & $4$B & $0.836$ & $0.233$ & -  & $0.804$ &- & -  & -  & - 
             \\
            \textcolor{mgt_blue}{$\bullet$}~PixWizard~\cite{lin2024pixwizard} & ICLR'25 & $2$B & $0.845$ & $0.248$ & $\mathbf{0.069}$ & $0.798$ & $0.884$ & $0.265$ & $0.063$ & $0.876$ 
            \\
            \textcolor{mgt_blue}{$\bullet$}~UniReal~\cite{chen2024UniReal} & CVPR'25 & $5$B & $0.851$ & $0.285$ & $0.099$ & $0.790$ & \underline{$0.903$} & $\mathbf{0.308}$ & $0.081$ & $0.837$
            \\
            \textcolor{mgt_blue}{$\bullet$}~GoT~\cite{fang2025got} &  NeurIPS'25 &  $6$B & $0.864$ & $0.276$ &- & - & - & - &-&- 
            \\
            \textcolor{mgt_blue}{$\bullet$}~OminiGen2~\cite{wu2025omnigen2} & arXiv'25 & $7$B & \underline{$0.876$} & $\mathbf{0.309}$ &- & $0.822$ & - & - & - &- 
            \\
            \textcolor{mgt_blue}{$\bullet$}~EditAR~\cite{mu2025editar} & ICLR'25 & $3$B &  - & - &-&-& $0.867$ & -& $0.103$ & $0.804$
            \\
            \textcolor{mgt_blue}{$\bullet$}~NEP~\cite{wu2025nep} & arXiv'25  & $3$B & $0.871$ & $0.307$ & $0.078$ & $\mathbf{0.844}$ &-&-&-&-
            \\
            \textcolor{mgt_blue}{$\bullet$}~VAREdit~\cite{varedit2025} & arXiv'25 & $8$B& \underline{$0.876$} & $0.280$ & $0.094$ & $0.825$ & $0.901$ & $0.287$ & $0.083$ & $0.844$
            \\
            \midrule
            \textcolor{mgt_dark}{$\bullet$}~\textbf{\textsc{\textcolor{mgt_gray}{Edit}\textcolor{mgt_red}{M}\textcolor{mgt_gray}{G}\textcolor{mgt_blue}{T}}} & \textbf{Ours} & $1$B & \cellcolor{my_red!7}$\mathbf{0.878}$ & \cellcolor{my_red!7}\underline{$0.308$} & \cellcolor{my_red!7}$0.093$ & \cellcolor{my_red!7}\underline{$0.832$} & \cellcolor{my_red!7}\textbf{$0.911$} & \cellcolor{my_red!7}\underline{$0.301$}& \cellcolor{my_red!7}$0.058$ & \cellcolor{my_red!7}$\mathbf{0.881}$
            \\
            \bottomrule
        \end{tabular}
    }
\end{table*}

%% file: sec/3_experiments.tex
\section{Experiments}
\label{sec:experiments}

We evaluate \name{} on standard benchmarks: Emu Edit~\cite{sheynin2024emu}, MagicBrush~\cite{zhang2024magicbrush}, AnyBench~\cite{yu2025anyedit}, and GEdit-EN-full~\cite{liu2025step1x}. Details are available in the main paper~\cite{chow2026editmgt}.

\noindent\textbf{Quantitative Results.}
As reported in Table~\ref{tab:eval_image_editing}, EditMGT achieves state-of-the-art CLIP$_{\text{im}}$ scores on both Emu Edit ($0.878$) and MagicBrush ($0.911$), with a $1.1\%$ improvement over the next best method on MagicBrush. DINO scores are state-of-the-art on MagicBrush and second-best on Emu Edit, while instruction adherence (CLIP$_{\text{out}}$) remains consistently competitive. On the GPT-based GEdit-EN-full benchmark~\cite{liu2025step1x}, EditMGT achieves competitive performance with FluxKontext.dev ($12$B) and surpasses VAREdit-$8$B, GoT-$6$B, and OminiGen2-$7$B, with notable margins of $+9.8\%$ on color alteration and $+17.6\%$ on style transfer. These results confirm that the MGT's localized decoding paradigm effectively prevents modifications from propagating into non-target regions.

\noindent\textbf{Qualitative Results \& Efficiency.}
Fig.~\ref{fig:quality} compares EditMGT against UltraEdit~(SD3), GoT~(6B), OminiGen2~(7B), and VAREdit~(8B). Key advantages include superior instruction comprehension (correctly reducing warm tones rather than increasing yellow) and faithful structural preservation (maintaining subject pose during style transfer). At $1024$$\times$$1024$ resolution, EditMGT completes an edit in ${\sim}$$2$ seconds with $13.8$ GB memory, $6\times$ faster than models of comparable quality.

%% file: sec/4_conclusion.tex
\section{Conclusion}
\label{sec:conclusion}

We presented \name{}, the first MGT-based framework that resolves the edit leakage problem inherent in diffusion models by leveraging localized token prediction. Multi-layer attention consolidation provides precise, mask-free edit localization, while region-hold sampling preserves non-target content during decoding. With $960$M parameters and $2$-second inference, EditMGT achieves state-of-the-art similarity across multiple benchmarks and competitive quality with models up to $12\times$ larger, establishing MGTs as an efficient paradigm for interactive image editing.